\documentclass[runningheads]{llncs}

\usepackage{accv}

\usepackage[T1]{fontenc}
\usepackage{accvabbrv}

\usepackage{graphicx,enumitem}
\usepackage{booktabs}
\usepackage{mathtools}
\usepackage{amssymb}
\usepackage{multirow}
\usepackage{makecell}
\usepackage[flushleft]{threeparttable}
\usepackage{tabularx}
\usepackage{xcolor}
\usepackage{textcomp}
\usepackage{pifont}
\usepackage{verbatim}
\usepackage{arydshln}

\usepackage[accsupp]{axessibility}

\usepackage{hyperref}

\usepackage{orcidlink}
\begin{document}

\title{vSV-ViT: Variable-size SuperVertex Vision Transformer for Cortical Surface Learning in Alzheimer's Disease}
\titlerunning{vSV-ViT: Variable-size SuperVertex Vision Transformer}


\author{
Geonwoo Baek\inst{1}\orcidID{0009-0001-2338-4575} \and
Ikbeom Jang\inst{1}\thanks{Corresponding author.}\orcidID{0000-0002-6901-983X} 
}

\authorrunning{G.~Baek et al.}

\institute{Department of Computer Science and Engineering,
Hankuk University of Foreign Studies, Seoul, Republic of Korea\\
\email{\{phlox3959, ijang\}@hufs.ac.kr}
}

\maketitle

\begin{abstract} 
Learning from 3D meshes is challenging because the data reside on non-Euclidean surfaces embedded in 3D space. This is particularly evident in domains such as brain cortical surface analysis, where existing models typically rely on ROI-agnostic, face-based, or fixed-size patches. Such patches can duplicate boundary vertices, conflate anatomically distinct regions, or incorporate non-cortical vertices such as those of the medial wall. We develop variable-size supervertex (vSV) partitioning, which groups cortical surface vertices into vSV patches similar to superpixels in images. Building on these vSVs, we design Variable-size SuperVertex Vision Transformer (vSV-ViT), a Vision Transformer that processes vSVs through padding, mask-aware patch embedding, and vertex position embedding. We demonstrate the benefits of the proposed framework through a quantitative analysis of vertex assignment in surface partitioning and downstream tasks using Alzheimer’s disease brain image datasets. For downstream tasks, we use meshes of cortical thickness and cortical curvature as input to various encoder models and evaluate classification performance. The superior predictive accuracy of vSV-ViT over recent surface-based models supports its effectiveness for cortical surface learning, with AD-related and sex prediction serving as downstream validation. Extending the framework to other cortical surface applications and broader 3D mesh domains remains future work. Code is available at \url{https://github.com/labhai/vSV-ViT}.



\end{abstract}

\section{Introduction}
Self-attention enables transformers to model long-range dependencies, but applying attention at the pixel level is computationally prohibitive. Vision Transformer (ViT)~\cite{dosovitskiy2021vit} addresses this by partitioning an image into patches and applying attention to patch tokens, offering a practical trade-off between representational capacity and computational burden. This patch-based formulation has also motivated the adaptation of ViTs to non-Euclidean data such as brain cortical surfaces.



Alzheimer's disease (AD) is characterized by neuropathology that accumulates over time, and confirmation often relies on $\beta$-amyloid (A$\beta$) and tau biomarkers measured by positron emission tomography (PET) or in cerebrospinal fluid (CSF). However, PET and CSF are limited in accessibility due to cost and invasiveness. MRI-based prediction of PET/CSF-derived pathology can support prescreening prior to testing~\cite{jang2024mssmab}. Beyond prediction, MRI-based approaches have also been extended to generative tasks, including longitudinal modeling of AD progression~\cite{lee2027adpdit} and PET image synthesis~\cite{moon2026cyclic}. The cerebral cortex, the outer layer of the brain, is particularly relevant to AD because disease-related neurodegeneration is reflected in cortical atrophy and morphological alterations~\cite{jang2022mssm}. This motivates surface-based learning on template-registered meshes, in which corresponding cortical locations align across subjects, as explored by recent models such as the Surface Vision Transformer (SiT)~\cite{dahan2022sit}. In this study, using cortical thickness (CT) and curvature (Curv) as inputs, we evaluate three AD-related prediction tasks and a sex prediction task: multiclass clinical diagnosis (Dx) among cognitively normal (CN), mild cognitive impairment (MCI), and AD groups, binary PET-derived A$\beta$ and tau positivity, and sex.

SiT~\cite{dahan2022sit} and MS-SiT~\cite{dahan2024mssit} demonstrated that ViT-style tokenization can be applied to registered cortical surfaces, enabling long-range attention over surface patches. However, regular icospheric, face-based, fixed-size patching is not anatomy-aware: patches can cross ROI boundaries, duplicate boundary vertices, and include non-cortical medial-wall vertices depending on token construction. SV-ViT~\cite{baek2026svvit} addressed part of this limitation by introducing ROI-informed supervertices, which group vertices similarly to superpixels in images, thereby alleviating ROI mixing. Nevertheless, its face-based, fixed-size supervertex design can still share boundary vertices across adjacent tokens and leave cortical vertices unassigned or under-covered.

To address these issues, we propose an ROI-preserving, vertex-based cortical surface partitioning that produces non-overlapping patches without ROI mixing or boundary-vertex duplication. The resulting variable-size supervertices (vSVs) naturally vary in size across ROIs. We then introduce Variable-size SuperVertex Vision Transformer (vSV-ViT) that builds on supervertex-based surface transformer learning and supports variable-size cortical patches through padding, mask-aware patch embedding, and vertex position embedding (vtxPE). By enabling attention over ROI-pure, non-overlapping patches, our framework balances attention granularity and computational burden, thereby improving the modeling of cortical features for Dx, A$\beta$, tau, and sex.

Our primary contributions are as follows:
\begin{itemize}[
    label=\textbullet,
    leftmargin=1.5em,
    labelsep=0.5em,
    topsep=0pt,
]
\item We develop a vertex-based cortical surface partitioning method that preserves ROI boundaries, excludes non-cortical regions, and allows variable-size patches to avoid duplicated or unassigned vertices, called variable-size supervertex (vSV) partitioning.
\item We integrate variable-size supervertices into a ViT using padding and vertex masking for cortical surface learning, referred to as vSV-ViT.
\item We introduce a learnable positional embedding that enables the model to capture relationships among vertices within each patch, termed vertex position embedding (vtxPE).
\end{itemize}



\section{Related Work}
Cortical surface learning has been studied with graph- and mesh-based neural networks, including DiffusionNet~\cite{sharp2022diffusionnet} and SurfGNN~\cite{li2026surfgnn}. These methods operate directly on surface geometry or graph connectivity, but do not explicitly formulate ViT-style anatomical patch tokenization.

Surface Vision Transformer (SiT)~\cite{dahan2022sit} and MS-SiT~\cite{dahan2024mssit} adapt ViT-style patch tokens to spherical cortical representations, showing that Transformer attention can be applied to registered cortical meshes. Their patch construction is regular and icospheric, whereas our focus is anatomy-aware surface tokenization.

SV-ViT~\cite{baek2026svvit} is the closest prior work to ours, as it introduced ROI-informed supervertices for anatomically informed surface tokens. Our method retains this supervertex-based motivation but changes the cortical surface partitioning and model architecture: vSVs are vertex-based, non-overlapping, and variable in size, and vSV-ViT embeds them through padding, vertex masks, and vtxPE.


\section{Method}

\subsection{Variable-Size Supervertex Partitioning}
\begin{figure}[tb]
  \centering
  \includegraphics[width=\textwidth]{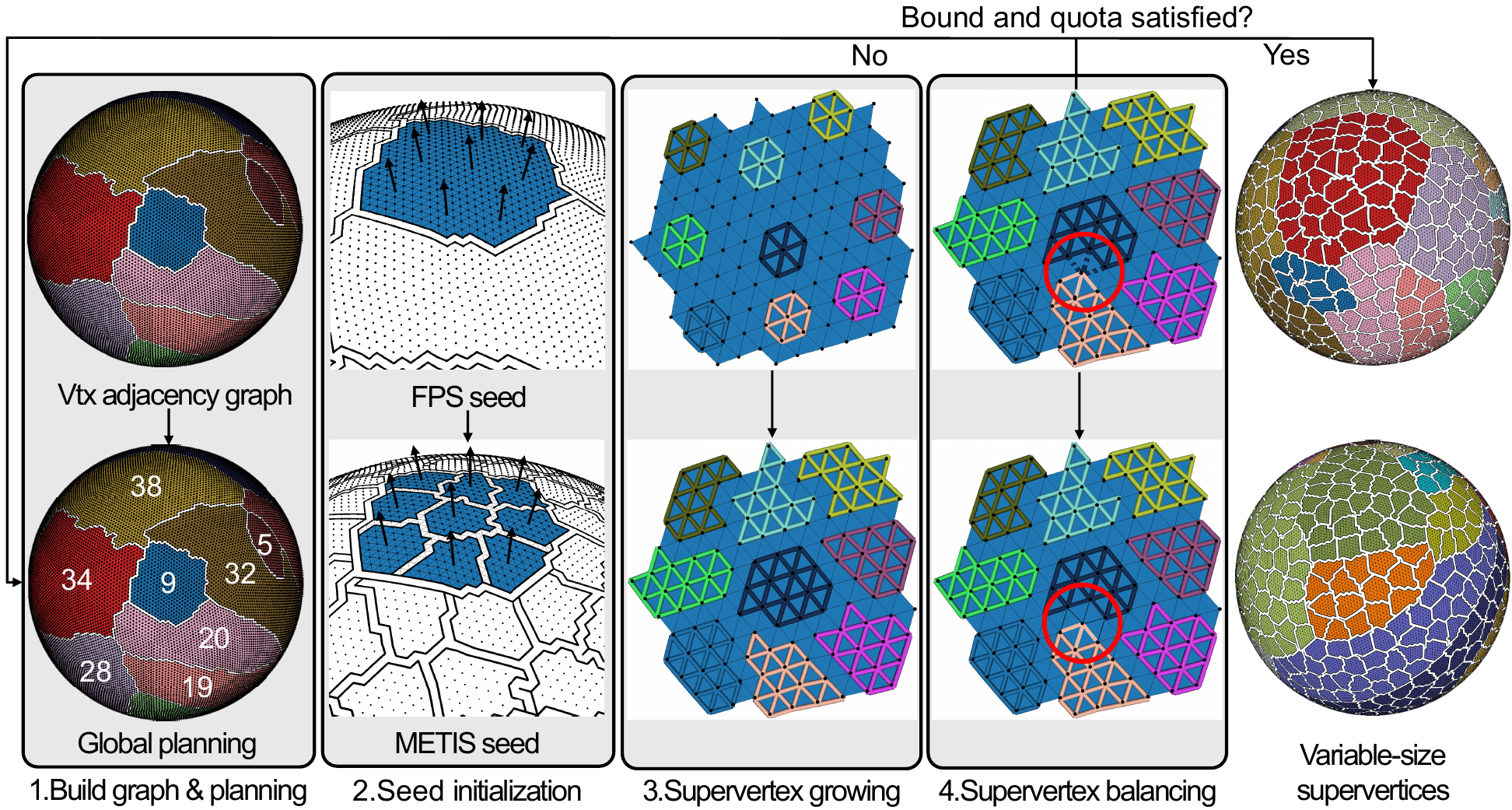}
  \caption{Overview of the proposed variable-size supervertex (vSV) partitioning pipeline. vSV partitioning is a vertex-based cortical surface partitioning method that groups vertices into variable-size supervertices within ROI boundaries to avoid ROI mixing, vertex duplication, and cortical vertex under-coverage.}
  \label{fig1}
\end{figure}

\cref{fig1} illustrates the overall partitioning pipeline. We register FreeSurfer cortical surfaces to fsaverage6 (ico6; 40,962 vertices per hemisphere) and partition each hemisphere into ROI-constrained variable-size supervertices (vSVs) using the Desikan--Killiany atlas~\cite{desikan2006automated}. To mitigate ROI mixing caused by atlas artifacts, minor disconnected fragments (<10\% of an ROI) are reassigned to the adjacent ROI with the largest boundary contact; non-cortical labels (e.g., the medial wall) are excluded. We model the ico6 mesh as a graph $G=(V,E)$ with 1-ring adjacency and construct connected vSVs within each ROI under a global size bound $L \le |SV| \le H$. The procedure comprises four stages: global planning, seed initialization, supervertex growth, and supervertex balancing, with fallback loops when constraints are not met.

\noindent\textbf{Global planning.}
We set the target number of vSVs to $K_{\text{total}}=642$ per hemisphere, corresponding to the number of vertices at ico3 resolution, to balance surface representation granularity and computational burden. Larger vSVs blur intra-patch detail; smaller ones increase the number of tokens and FLOPs. We then search for a feasible bound $(L,H)$ and ROI-wise counts $\{K_r\}$ such that $\sum_r K_r = K_{\text{total}}$ and each ROI $r$ (with $n_r$ vertices) satisfies:
\begin{equation}
\left\lceil \frac{n_r}{H}\right\rceil \le K_r \le \left\lfloor \frac{n_r}{L}\right\rfloor.
\end{equation}
Among feasible allocations, we select $\{K_r\}$ to reduce ROI-wise imbalance by minimizing $\sum_r (n_r/K_r-\bar{s})^2$, where $\bar{s}=(\sum_r n_r)/K_{\text{total}}$. We try candidates from tighter to looser bounds and proceed only if later stages succeed.

\noindent\textbf{Seed initialization.}
For each ROI, we handle multiple connected components by distributing $K_r$ among them as needed. We initialize $K_r$ seeds via farthest-point sampling (FPS) on vertex direction vectors, and optionally refine the seeds using METIS~\cite{karypis1998metis} to obtain balanced subgraphs (falling back to FPS if METIS fails).

\noindent\textbf{Supervertex growing.}
Within each ROI component, we define near-equal integer quotas (size difference $\le 1$) and grow SVs by absorbing unassigned 1-ring boundary vertices, which preserves connectivity. SVs with lower quota fill ratios are expanded first. If any SV fails to meet its quota after growing, we update seeds based on the current assignment and repeat the seed--grow step up to a fixed number of retries.

\noindent\textbf{Supervertex balancing.}
We then enforce the bound $L \le |SV| \le H$ by transferring boundary vertices between adjacent SVs on an SV adjacency graph while preserving connectivity. If any SV still violates the bound after balancing, we reject the current plan and return to \textbf{global planning} to retry with a looser $(L,H,\{K_r\})$.

Finally, we obtain 642 ROI-preserving vSVs per hemisphere (cortex-only, fully-covering, non-overlapping). Within each vSV, vertices are ordered by their global vertex indices. Each vSV is represented by its vertex-index list and padded with $-1$ to the maximum vSV size ($V_{\max}=69$) for model input.

\subsection{Variable-size SuperVertex Vision Transformer}
\begin{figure}[tb]
  \centering
  \includegraphics[width=\textwidth]{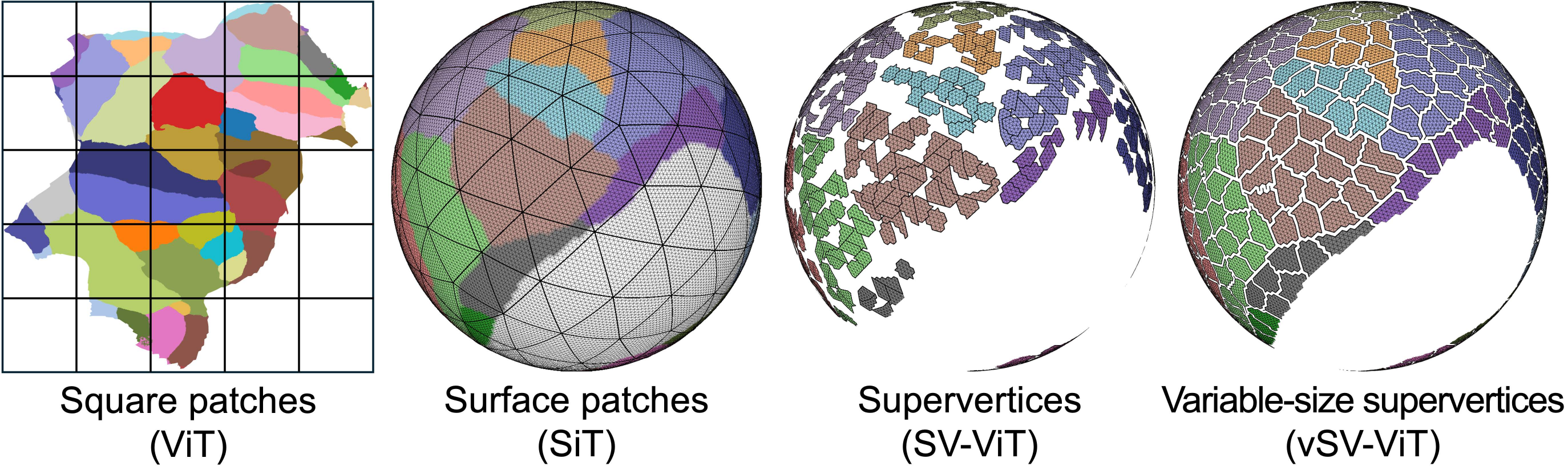}
\caption{Differences in patch partitioning methods among ViT, SiT, SV-ViT, and vSV-ViT. SV-ViT employs fixed-size, face-based partitioning. vSV-ViT adopts variable-size, vertex-based partitioning to avoid unassigned or duplicated vertices.} \label{fig2}
\end{figure}

vSV-ViT represents the cortical surface by vSVs. \cref{fig2} shows the differences in patch partitioning methods among ViT, SiT, SV-ViT, and vSV-ViT. ViT partitions a 2D image into fixed-size square patches, whereas SiT partitions the cortical surface into regular, equal-size, face-based surface patches. SV-ViT further constrains the patches by anatomical ROIs to prevent ROI mixing. However, maintaining fixed-size, face-based patches under ROI constraints can leave cortical vertices unassigned and duplicate boundary vertices across adjacent patches. 
In contrast, vSV-ViT uses ROI-constrained, variable-size, vertex-based partitioning to avoid duplicated and unassigned cortical vertices.

\begin{figure}[tb]
  \centering
  \includegraphics[width=\textwidth]{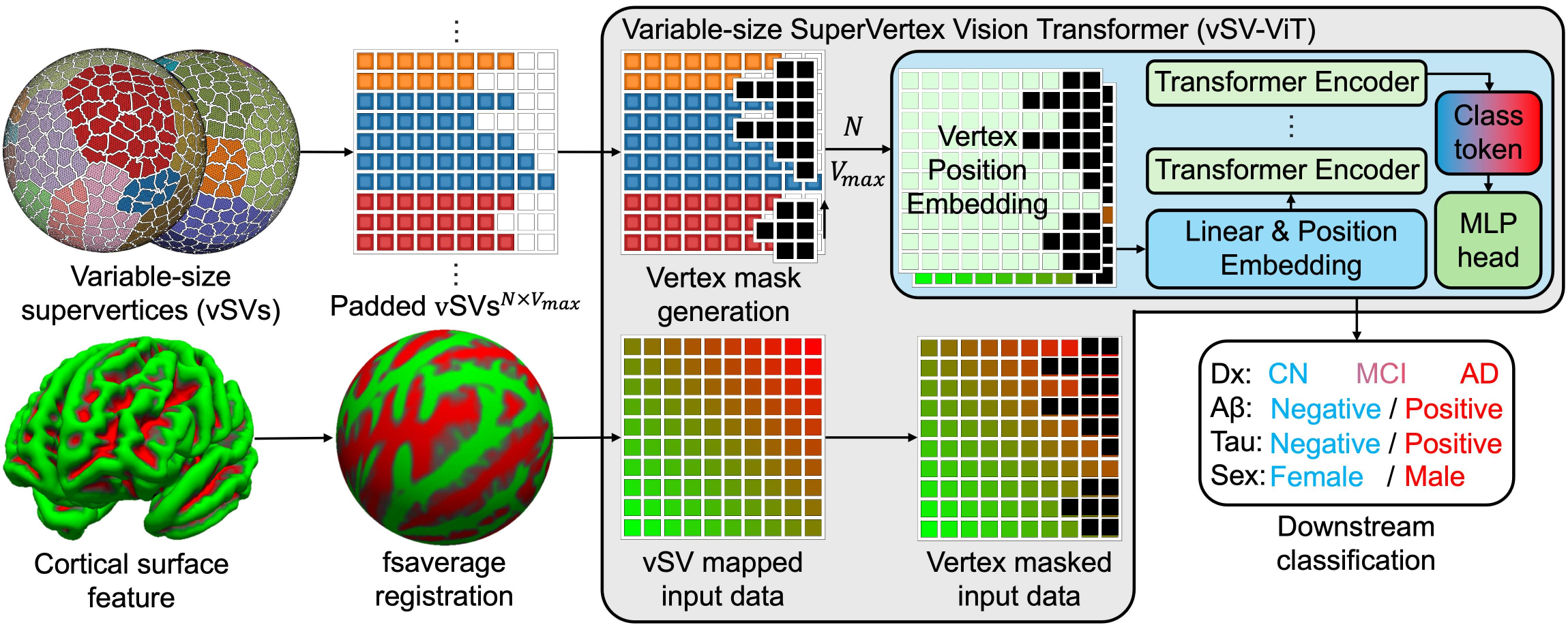}
\caption{Architecture of Variable-size SuperVertex Vision Transformer (vSV-ViT). Vertex position embedding learns the relationships among vertices within each vSV, whereas token-level position embedding encodes the relative arrangement of the vSV tokens and the class token. Vertex masking, padding, and mask-aware patch embedding enable vSV-ViT to process variable-size supervertices.}
\label{fig3}
\end{figure}

\cref{fig3} shows the vSV-ViT architecture. Given per-vertex cortical features on the ico6 mesh with $C$ channels and a precomputed vSV map, we collect the vertices of each vSV and pad them to a fixed maximum length $V_{\max}$. This yields an input tensor $\mathbf{X}\in\mathbb{R}^{B\times C\times N\times V_{\max}}$, where $N$ is the number of vSVs (both hemispheres). We build a binary mask $\mathbf{M}\in\{0,1\}^{N\times V_{\max}}$ from the vSV map and zero out padded entries before embedding.

\noindent\textbf{Vertex position embedding.}
Because each vSV contains multiple cortical vertices, collapsing the padded vertex list into a single token without positional information would limit the sensitivity of patch embedding to intra-vSV vertex locations. To let the model learn positional relationships among vertices within each vSV, we add a learnable vtxPE to the padded feature matrix:
\begin{equation}
\mathbf{z}_i = \mathbf{x}_i + \mathbf{p}_i,
\quad i=1,\dots,N,
\end{equation}
where $\mathbf{x}_i\in\mathbb{R}^{C\times V_{\max}}$ is the padded feature matrix of the $i$-th vSV and $\mathbf{p}_i\in\mathbb{R}^{C\times V_{\max}}$ denotes the learnable vtxPE for intra-vSV positional learning. The vtxPE slots follow the global vertex-index ordering within each vSV, providing a consistent positional convention independent of anatomical ordering.

\noindent\textbf{Mask-aware patch embedding.}
We apply the binary mask to the vtxPE-added vSV feature matrix, flatten the vertices within each vSV, and apply a linear embedding layer to obtain a single token per vSV:
\begin{equation}
\tilde{\mathbf{x}}_i
=
\mathbf{z}_i \odot \mathbf{m}_i,
\qquad
\mathbf{t}_i
=
\mathrm{Linear}\big(\mathrm{vec}(\tilde{\mathbf{x}}_i)\big),
\quad i=1,\dots,N,
\end{equation}
where $\mathbf{m}_i\in\{0,1\}^{1\times V_{\max}}$ is the vertex mask broadcast along the channel dimension, $\odot$ denotes element-wise multiplication, and $\mathrm{vec}(\cdot)$ concatenates the vertex and channel dimensions.

\noindent\textbf{Transformer encoder and classification head.}
The embedded vSV tokens are stacked as
$\mathbf{T}=[\mathbf{t}_1,\dots,\mathbf{t}_N]\in\mathbb{R}^{B\times N\times D}$,
where $D$ is the token embedding dimension. We prepend a learnable classification token to $\mathbf{T}$ and add a learnable token-level positional embedding to the resulting sequence. The sequence is then processed by a standard Transformer encoder composed of stacked multi-head self-attention and feed-forward blocks with residual connections and layer normalization. The output classification token is fed into a multi-layer perceptron (MLP) head for the corresponding classification task.

\subsection{Downstream Validation on Alzheimer's Disease Datasets}
We performed four classification tasks for Dx, A$\beta$ positivity, tau positivity, and sex to evaluate the proposed framework on a multi-cohort dataset. In addition to two cortical surface features, including CT and Curv, we provided vSV-ViT with the vSV partition map, which defines $N=1,284$ vSVs (both hemispheres) and a maximum vSV size of $V_{\max}{=}69$.

\section{Experiments}

\subsection{Datasets and Preprocessing}




We used public data from ADNI-GO/2/3/4 of the Alzheimer's Disease Neuroimaging Initiative (ADNI)~\cite{petersen2010adni} and OASIS-3 of the Open Access Series of Imaging Studies (OASIS)~\cite{lamontagne2019oasis3}. To maximize the number of eligible participants while reducing protocol heterogeneity, we restricted MRI inputs to accelerated scans acquired on Siemens 3T scanners in both cohorts. To increase cohort diversity while controlling scanner heterogeneity, we pooled eligible ADNI and OASIS participants under the same Siemens 3T MRI-input restriction. We used T1-weighted (T1w) MR images to extract cortical features. Information on participants is presented in \cref{tab:demo}. Each participant contributed at most once to each task. For the Dx task, the first available quality-controlled T1w scan was used when multiple sessions were available. For A$\beta$ and tau, we selected the baseline PET measure and matched it to the closest quality-controlled T1w scan acquired within 365 days. Because not all subjects had a matched PET scan, the participant counts differ across the Dx, A$\beta$, and tau classifications. A$\beta$ and tau labels were derived from cohort-provided PET measures (AV45~\cite{johnson2013florbetapir} for A$\beta$ and AV1451~\cite{marquie2015av1451} for tau) using predefined cohort-specific positivity criteria. In ADNI, A$\beta$ positivity was defined using a whole cerebellum normalized summary standardized uptake value ratio (SUVR) cutoff of 1.11~\cite{royse2021amyloid_thresholds}, and tau positivity was defined using a meta-ROI SUVR cutoff of 1.37~\cite{meyer2020biomarker}. In OASIS, A$\beta$ positivity was defined using a Centiloid cutoff of 20.6 derived from the regional spread function-corrected mean cortical SUVR, and tau positivity was defined using a tauopathy cutoff of 1.22~\cite{lamontagne2019oasis3}. For Dx, we considered a multiclass classification setting comprising cognitively normal (CN), mild cognitive impairment (MCI), and AD.

All T1w MR images were processed with FreeSurfer v8.1.0~\cite{dale1999cortical,fischl2004,fischl2012freesurfer} to reconstruct cortical surfaces and extract vertex-wise CT and Curv. Only data that passed FreeSurfer quality control were retained. The data were then registered to the fsaverage template, represented as an icosahedral subdivision level 6 (40,962 vertices per hemisphere), and left/right hemispheres were concatenated.

\begin{table}[tb]
  \setlength{\tabcolsep}{4pt}
  \small
  \centering
  \caption{Participant statistics of the datasets in this study. Participants from ADNI and OASIS were pooled for each target. For A$\beta$ and tau, status indicates PET-based negativity or positivity; for diagnosis (Dx), status indicates CN, MCI, or AD. Values are the number of subjects (n), within-target proportion (Prop.), percentage of females (\%F), and age (mean $\pm$ std (min--max)). No subject was included more than once within each target.}
  \label{tab:demo}
  \begin{tabularx}{\textwidth}{l
  >{\centering\arraybackslash}p{1.5cm}
  >{\centering\arraybackslash}p{1.25cm}
  >{\centering\arraybackslash}p{1.5cm}
  >{\centering\arraybackslash}p{1.25cm}
  >{\centering\arraybackslash}X
  }
  \toprule
  Target & Status & n & Prop. (\%) & \%F & Age \\
  \midrule
  \multirow{2}{*}{Amyloid}
  & Negative & 814 & 57.9 & 54.4 & 69.76\textpm7.97 (42.70--95.45) \\
  & Positive & 592 & 42.1 & 54.2 & 74.09\textpm7.36 (55.12--90.35) \\
  \cmidrule(lr){1-6}

  \multirow{2}{*}{Tau}
  & Negative & 814 & 73.6 & 55.4 & 70.69\textpm8.35 (46.24--94.41) \\
  & Positive & 292 & 26.4 & 61.0 & 72.63\textpm7.58 (46.44--89.52) \\
  \cmidrule(lr){1-6}

  \multirow{3}{*}{Diagnosis}
  & CN  & 1681 & 60.5 & 59.8 & 69.50\textpm8.44 (42.69--95.64) \\
  & MCI & 650  & 23.4 & 49.2 & 72.40\textpm7.79 (55.04--97.45) \\
  & AD  & 447  & 16.1 & 45.6 & 75.55\textpm8.13 (50.79--97.08) \\
  \cmidrule(lr){1-6}

  \multirow{2}{*}{Sex}
  & Female & 1529 & 55.0 & 100.0 & 70.12\textpm8.66 (43.52--95.97) \\
  & Male   & 1249 & 45.0 & 0.0   & 72.42\textpm8.22 (42.69--97.45) \\
  \bottomrule
  \end{tabularx}
\end{table}

\subsection{Experimental Setting}
We evaluated the proposed framework on four classification tasks: multiclass Dx, A$\beta$ positivity, tau positivity, and sex. Because tau pathology is often more tightly coupled with neurodegeneration than A$\beta$, tau positivity is frequently expected to be more predictable from structural MRI-derived morphometry. Nevertheless, MRI-based biomarker prediction performance depends on cohort composition, label definitions, and prevalence, and amyloid prediction can match or exceed tau prediction; for example, a prior study reported higher area under the receiver operating characteristic curve (AUROC) for predicting PET-determined A$\beta$ status than tau status using MRI volumes and readily available clinical variables~\cite{lew2023atn}.

We reported AUROC as the primary metric and area under the precision–recall curve (AUPRC) as the secondary metric across all downstream experiments. For multiclass Dx, both AUROC and AUPRC were computed using macro-averaged one-vs-rest evaluation. We compared our method against recent models that can process the cortical surface, including SiT~\cite{dahan2022sit}, DiffusionNet~\cite{sharp2022diffusionnet}, MS-SiT~\cite{dahan2024mssit}, SurfGNN~\cite{li2026surfgnn}, and SV-ViT~\cite{baek2026svvit}. We selected these methods based on their ability to process cortical surface data, rather than restricting the comparison to methods originally developed for the specific Dx, A$\beta$, tau, sex prediction tasks.

We used 4-fold cross-validation with a fixed fold assignment so that all models were trained and tested on identical data and splits. For each fold, three folds were used for training, and one for testing; a validation set was created by holding out 10\% of the training data with stratification. We used the AdamW optimizer and a cosine annealing learning rate scheduler for all models. To address class imbalance, we optimized a class-weighted cross-entropy (CE) loss for Dx and a class-weighted binary cross-entropy (BCE) loss for A$\beta$, tau, and sex. Unless otherwise stated, we reported the mean $\pm$ standard deviation (std) across folds. Statistical significance was assessed on pooled out-of-fold test predictions using two-sided paired DeLong tests for binary AUROC and two-sided paired permutation tests with 10,000 permutations for multiclass macro-AUROC and all AUPRC comparisons. P-values were adjusted using the Holm method across the five model comparisons within each feature–target–metric combination.

For external validation, we trained vSV-ViT on OASIS and independently evaluated it on ADNI. Within OASIS, we used 4-fold cross-validation, using three folds for training and one fold for validation in each iteration.

\subsection{Surface Partitioning Analysis}
\noindent\textbf{Comparison of surface partitioning methods.}

SiT-style triangular-patch partitioning~\cite{dahan2022sit} defines patches according to a regular icospheric grid rather than anatomical ROI boundaries. As a result, a patch can span multiple ROIs, and boundary vertices can be included in multiple patch tokens when triangular faces are converted into vertex-indexed inputs. SV-ViT~\cite{baek2026svvit} alleviates ROI mixing by using ROI-informed supervertices. However, because SV-ViT uses face-based SVs, boundary vertices can be shared by adjacent SV tokens. Moreover, enforcing fixed-size SVs can leave cortical vertices unassigned. In contrast, vSV partitioning constructs cortex-only, vertex-based, non-overlapping vSVs within ROI boundaries. \cref{tab:partitioning_comparison} presents a quantitative analysis of cortical surface partitioning.


\begin{table}[tb]
    \centering
    \setlength{\tabcolsep}{5.0pt}
    \renewcommand\bfdefault{b}
    \begin{threeparttable}
        \centering
        \small
        \caption{Quantitative analysis of vertex assignment in surface partitioning. Values are averaged over the left and right hemispheres. Duplicated vertices denote vertices assigned to multiple token inputs, and unassigned vertices denote retained cortical vertices not covered by any token input. ROI-mixed patches denote patches containing vertices from more than one ROI. Medial-wall vertices denote medial-wall vertices included in token inputs.
}
        \label{tab:partitioning_comparison}
        \begin{tabularx}{\textwidth}{@{\hspace{8pt}}l
        >{\centering\arraybackslash}p{2cm}
        >{\centering\arraybackslash}p{2cm}
        >{\centering\arraybackslash}p{2cm}
        >{\centering\arraybackslash}p{2.2cm}
        }
            \toprule
            \raisebox{-1.15\baselineskip}{Model} &
            \shortstack{Duplicated\\vertices (\%)} &
            \shortstack{Unassigned\\vertices (\%)} &
            \shortstack{ROI-mixed\\patches (\%)} &
            \shortstack{Medial wall\\vertices (\%)} \\
            \midrule
            SiT~\cite{dahan2022sit} &
            18.0 &
            0 &
            73.3 &
            10.1 \\
            SV-ViT~\cite{baek2026svvit} &
            21.1 &
            27.4 &
            0 &
            0 \\
            vSV-ViT (ours) &
            0 &
            0 &
            0 &
            0 \\
            \bottomrule
        \end{tabularx}
    \end{threeparttable}
\end{table}

\subsection{vSV-ViT against recent models}

\noindent\textbf{Classification Performance}

\cref{tab:main_results} summarizes the classification performance of baseline models applicable to the cortical surface (SiT, DiffusionNet, MS-SiT, SurfGNN, and SV-ViT) and our vSV-ViT on Dx, A$\beta$, tau, and sex using CT and Curv as inputs. Overall, vSV-ViT achieves higher classification performance than recent surface-based models.

Among the compared models, DiffusionNet and SurfGNN operate on surface geometry or graph connectivity without explicitly imposing ROI-preserving patch tokens. SiT and MS-SiT use regular icospheric, face-based patches, which can cross ROI boundaries and include duplicated or medial-wall vertices depending on the vertex-indexed input construction. SV-ViT improves this design by introducing ROI-informed supervertices and therefore alleviates ROI mixing, but it still follows a face-based, fixed-size tokenization scheme. In contrast, vSV-ViT uses vertex-based, non-overlapping, variable-size vSVs, so it simultaneously preserves ROI boundaries, keeps cortical vertices disjoint, avoids cortical vertex under-coverage, and excludes non-cortical medial-wall vertices during tokenization. Together, the partitioning statistics and ablation results suggest that the proposed vSV tokenization and mask-aware input framework improves cortical surface learning.

We observe that CT generally provides stronger predictive signals than Curv for Dx, A$\beta$, and tau. Notably, vSV-ViT improves performance across both feature sets, indicating its effectiveness at learning from cortical surface features.

\begin{table}[!tb]
    \centering
    \setlength{\tabcolsep}{1.8pt}
    \newcommand{\pstars}[1]{\rlap{\textsuperscript{#1}}}
    \renewcommand\bfdefault{b}
    \begin{threeparttable}
        \centering
        \small
        \caption{Classification performance of vSV-ViT against recent models that can process the cortical surface. Diagnosis (Dx) is evaluated as a multi-class classification among CN, MCI, and AD; A$\beta$ and tau are evaluated as binary classifications of PET-based positivity status; and sex is evaluated as a binary classification. Cortical thickness and cortical curvature are evaluated separately as inputs. We report mean $\pm$ std. We highlight the \textbf{best} and \underline{second-best} results. Statistical significance is based on Holm-adjusted $p$-values: * $p<0.05$; ** $p<0.01$ compared with vSV-ViT.}
        \label{tab:main_results}
        \begin{tabularx}{\textwidth}{@{\hspace{3pt}}lc
        >{\centering\arraybackslash}p{2.0cm}
        >{\centering\arraybackslash}p{2.0cm}
        >{\centering\arraybackslash}p{2.0cm}
        >{\centering\arraybackslash}p{2.0cm}
        }
            \toprule
            \phantom{0} & Target & \multicolumn{2}{c}{Cortical thickness} & \multicolumn{2}{c}{Cortical curvature} \\
            \cmidrule(lr){3-4}
            \cmidrule(lr){5-6}
             Model &   & AUROC & AUPRC & AUROC & AUPRC \\
             \midrule
             SiT~\cite{dahan2022sit} & Dx &
             0.629\textpm0.014\pstars{**} &
             0.442\textpm0.013\pstars{**} &
             0.615\textpm0.037\pstars{*} &
             0.419\textpm0.022\pstars{*} \\
             DiffusionNet~\cite{sharp2022diffusionnet} &  &
             0.662\textpm0.020\pstars{*} &
             0.486\textpm0.025\pstars{*} &
             0.625\textpm0.028\pstars{**} &
             0.423\textpm0.036\pstars{**} \\
             MS-SiT~\cite{dahan2024mssit} & \phantom{0} &
             \underline{0.707\textpm0.016} &
             \underline{0.539\textpm0.013}\pstars{*} &
             0.634\textpm0.020 &
             0.447\textpm0.024 \\
             SurfGNN~\cite{li2026surfgnn} & \phantom{0} &
             0.657\textpm0.015\pstars{**} &
             0.470\textpm0.017\pstars{**} &
             \underline{0.657\textpm0.019} &
             \textbf{0.465\textpm0.022} \\
             SV-ViT~\cite{baek2026svvit} & \phantom{0} &
             0.606\textpm0.027\pstars{**} &
             0.423\textpm0.025\pstars{**} &
             0.633\textpm0.018 &
             0.439\textpm0.013 \\
             vSV-ViT (ours) & \phantom{0} &
             \textbf{0.709\textpm0.022} &
             \textbf{0.541\textpm0.026} &
             \textbf{0.664\textpm0.009} &
             \underline{0.465\textpm0.009} \\
             \midrule
             SiT~\cite{dahan2022sit} & A$\beta$ &
             0.602\textpm0.007\pstars{**} &
             0.526\textpm0.022\pstars{**} &
             0.531\textpm0.017\pstars{**} &
             0.456\textpm0.028\pstars{**} \\
             DiffusionNet~\cite{sharp2022diffusionnet} &  &
             0.626\textpm0.028\pstars{**} &
             0.560\textpm0.046\pstars{**} &
             0.536\textpm0.022\pstars{**} &
             0.449\textpm0.033\pstars{**} \\
             MS-SiT~\cite{dahan2024mssit} & \phantom{0} &
             \underline{0.646\textpm0.014} &
             \underline{0.582\textpm0.029} &
             \underline{0.562\textpm0.023}\pstars{**} &
             0.464\textpm0.022\pstars{**} \\
             SurfGNN~\cite{li2026surfgnn} & \phantom{0} &
             0.621\textpm0.033 &
             0.550\textpm0.052 &
             0.544\textpm0.069\pstars{**} &
             \underline{0.472\textpm0.060}\pstars{*} \\
             SV-ViT~\cite{baek2026svvit} & \phantom{0} &
             0.580\textpm0.031\pstars{**} &
             0.494\textpm0.023\pstars{**} &
             0.542\textpm0.025\pstars{**} &
             0.451\textpm0.035\pstars{**} \\
             vSV-ViT (ours) & \phantom{0} &
             \textbf{0.648\textpm0.018} &
             \textbf{0.594\textpm0.015} &
             \textbf{0.619\textpm0.032} &
             \textbf{0.552\textpm0.049} \\
             \midrule
             SiT~\cite{dahan2022sit} & Tau &
             0.629\textpm0.033 &
             0.386\textpm0.045 &
             0.560\textpm0.031 &
             0.322\textpm0.041 \\
             DiffusionNet~\cite{sharp2022diffusionnet} &   &
             0.700\textpm0.031\pstars{*} &
             0.442\textpm0.060 &
             0.527\textpm0.044 &
             0.301\textpm0.042 \\
             MS-SiT~\cite{dahan2024mssit} & \phantom{0} &
             0.696\textpm0.044\pstars{*} &
             \underline{0.478\textpm0.053} &
             \underline{0.561\textpm0.035} &
             0.333\textpm0.059 \\
             SurfGNN~\cite{li2026surfgnn} & \phantom{0} &
             \underline{0.707\textpm0.020}\pstars{**} &
             0.453\textpm0.039 &
             0.537\textpm0.046 &
             0.296\textpm0.027 \\
             SV-ViT~\cite{baek2026svvit} & \phantom{0} &
             0.640\textpm0.032\pstars{*} &
             0.395\textpm0.055 &
             0.559\textpm0.040 &
             \underline{0.340\textpm0.031} \\
             vSV-ViT (ours) & \phantom{0} &
             \textbf{0.725\textpm0.018} &
             \textbf{0.493\textpm0.047} &
             \textbf{0.570\textpm0.018} &
             \textbf{0.343\textpm0.020} \\
             \midrule
             SiT~\cite{dahan2022sit} & Sex &
             0.788\textpm0.017\pstars{*} &
             0.754\textpm0.023\pstars{**} &
             0.737\textpm0.025\pstars{**} &
             0.680\textpm0.031\pstars{**} \\
             DiffusionNet~\cite{sharp2022diffusionnet} & &
             0.589\textpm0.022\pstars{**} &
             0.528\textpm0.031\pstars{**} &
             0.635\textpm0.028\pstars{**} &
             0.559\textpm0.028\pstars{**} \\
             MS-SiT~\cite{dahan2024mssit} & \phantom{0} &
             \underline{0.821\textpm0.104} &
             \underline{0.790\textpm0.131} &
             \underline{0.835\textpm0.014} &
             \underline{0.804\textpm0.015}\pstars{**} \\
             SurfGNN~\cite{li2026surfgnn} & \phantom{0} &
             0.531\textpm0.061\pstars{**} &
             0.485\textpm0.057\pstars{**} &
             0.682\textpm0.096\pstars{**} &
             0.626\textpm0.113\pstars{**} \\
             SV-ViT~\cite{baek2026svvit} & \phantom{0} &
             0.703\textpm0.040\pstars{**} &
             0.656\textpm0.043\pstars{**} &
             0.817\textpm0.013 &
             0.790\textpm0.021\pstars{*} \\
             vSV-ViT (ours) & \phantom{0} &
             \textbf{0.827\textpm0.011} &
             \textbf{0.798\textpm0.025} &
             \textbf{0.853\textpm0.015} &
             \textbf{0.830\textpm0.016} \\
             \bottomrule
        \end{tabularx}
    \end{threeparttable}
\end{table}

\noindent\textbf{Computational efficiency.}
\cref{tab:computational_efficiency} compares the representation size and computational cost of the evaluated models. Compared with SV-ViT, vSV-ViT reduces the number of tokens from 2,560 to 1,284, the representation size from 276,480 to 88,596, and FLOPs from 153.31G to 70.91G. Of the 88,596 padded input slots, 73,573 contain cortical vertices and the remaining 17\% correspond to padding. Among the ViT-based models, vSV-ViT therefore has the smallest representation size, although its FLOPs remain higher than those of SiT and MS-SiT. In particular, compared with MS-SiT, vSV-ViT uses 2.2$\times$ fewer parameters but 52$\times$ more FLOPs, indicating that the proposed representation is not designed to minimize computational cost. Inference times for the ViT-based models range from 3.0 to 8.3\,ms, with vSV-ViT requiring 7.1\,ms.

\begin{table}[t]
\centering
\scriptsize
\setlength{\tabcolsep}{3.9pt}
\renewcommand{\arraystretch}{0.85}
\caption{Computational cost of the evaluated models. Token counts and representation sizes cover both hemispheres. DiffusionNet and SurfGNN do not use token-based representations. Inference time is reported as the average time per sample.}
\label{tab:computational_efficiency}
\footnotesize
\begin{tabular}{lrrrrr}
\toprule
Method & Tokens & Representation size & Parameters & FLOPs & Inference time \\
\midrule
SiT     & 640    & 97,920  & 3.71M  & 7.10G   & 3.0\,ms \\
DiffusionNet &  N/A    & 81,924  & 3.62M  & 723.01G & 110.0\,ms \\
MS-SiT  & 40,960 & 245,760 & 40.52M & 1.36G   & 8.3\,ms \\
SurfGNN &  N/A        & 81,924  & 0.04M  & 4.05G   & 62.3\,ms \\
SV-ViT  & 2,560  & 276,480 & 15.22M & 153.31G & 5.7\,ms \\
vSV-ViT & 1,284  & 88,596  & 18.34M & 70.91G  & 7.1\,ms \\
\bottomrule
\end{tabular}
\end{table}

\subsection{Ablation Study}
\noindent\textbf{Partitioning ablation study.}  
\cref{tab:partitioning_results} shows the effect of atlas choice and the balancing step on the vSV size distribution. The mean vSV size is similar across atlases, but the Destrieux atlas yields larger size variability than the Desikan--Killiany atlas. The balancing step consistently reduces the standard deviation and narrows the minimum--maximum range for both atlases, indicating that it improves vSV size uniformity while preserving the variable-size partitioning scheme. 

\begin{table}[t]
    \centering
    \setlength{\tabcolsep}{6.0pt}
    \renewcommand\bfdefault{b}
    \begin{threeparttable}
        \centering
        \small
        \caption{Ablation study of vSV partitioning. The default vSV partitioning uses the Desikan--Killiany atlas to define cortical ROIs, and we compare it with Destrieux-based partitioning to assess the effect of atlas choice. We also evaluate the effect of the vSV balancing by comparing partitions generated with and without balancing. For each hemisphere (hemi), values denote mean $\pm$ std vSV size, with the minimum--maximum size in parentheses.}
        \label{tab:partitioning_results}
        \begin{tabularx}{\textwidth}{@{\hspace{3pt}}l
        >{\centering\arraybackslash}p{2.6cm}
        >{\centering\arraybackslash}X
        >{\centering\arraybackslash}X
        }
            \toprule
            Atlas & SV balancing & Left hemi & Right hemi \\
            \midrule
            Destrieux~\cite{destrieux2010automatic} &
            X &
            58.4\textpm3.9 (38--77) &
            58.4\textpm3.8 (35--69) \\
             &
            O &
            58.4\textpm3.3 (39--64) &
            58.4\textpm3.5 (36--64) \\
            Desikan--Killiany~\cite{desikan2006automated} &
            X &
            57.2\textpm2.3 (51--69) &
            57.4\textpm2.2 (48--60) \\
             &
            O &
            57.2\textpm1.7 (52--69) &
            57.4\textpm1.6 (49--60) \\
            \bottomrule
        \end{tabularx}
    \end{threeparttable}
\end{table}

\noindent\textbf{Incremental ablation study.}
\cref{tab:ablation_results} presents an incremental ablation study by adding each component to the SV-ViT baseline: vertex-based non-overlapping partitioning (Vtx), variable-size patches (vSVs) enabled by padding and a mask-aware embedding, supervertex balancing step (Bal.), and vtxPE for learning positional relationships among vertices within each vSV. Adding all components results in the vSV-ViT framework.

All components improve AUROC. The largest overall gain is observed when vertex-based partitioning is added. This design also facilitates the exclusion of non-cortical vertices (e.g., the medial wall). Adding variable-size patches further improves AUROC, consistent with avoiding vertex under-coverage that can occur when enforcing uniform patch sizes. Supervertex balancing and vtxPE provide additional gains. AUPRC also generally increases as the components are added, and the complete vSV-ViT framework outperforms the SV-ViT baseline across all tasks and input features. These results indicate that each component contributes to the overall performance gains.

\begin{table}[!t]
\centering
\setlength{\tabcolsep}{3.5pt}
\renewcommand\bfdefault{b}
\begin{threeparttable}
\caption{Incremental ablation study of vSV partitioning and vSV-ViT components. Diagnosis (Dx) is evaluated as multiclass classification of CN, MCI, and AD; A$\beta$ and tau are evaluated as binary PET-positivity classifications; and sex is evaluated as binary classification. Evaluated components are vertex-based partitioning (Vtx), variable-size supervertices with mask-aware padding (vSVs), supervertex balancing (Bal.), and vertex position embedding (vtxPE). The first and last rows correspond to SV-ViT~\cite{baek2026svvit} and vSV-ViT, respectively.}
\label{tab:ablation_results}
\small
\begin{tabularx}{\textwidth}{@{}
>{\centering\arraybackslash}p{0.65cm}
>{\centering\arraybackslash}p{0.65cm}
>{\centering\arraybackslash}p{0.7cm}
>{\centering\arraybackslash}p{0.7cm}
>{\centering\arraybackslash}p{0.8cm}
@{\hspace{9pt}}
>{\centering\arraybackslash}p{1.65cm}
>{\centering\arraybackslash}p{1.65cm}
>{\centering\arraybackslash}p{1.65cm}
>{\centering\arraybackslash}p{1.65cm}
@{}}
\toprule
Target &
\multicolumn{4}{c}{Components} &
\multicolumn{2}{c}{Cortical thickness} &
\multicolumn{2}{c}{Cortical curvature} \\
\cmidrule(lr{2pt}){2-5}
\cmidrule(lr){6-7}
\cmidrule(lr{1pt}){8-9}
& Vtx & vSVs & Bal. & VtxPE
& AUROC & AUPRC & AUROC & AUPRC \\
\midrule
Dx
& & & &
& 0.606$\pm$0.027 & 0.423$\pm$0.025
& 0.633$\pm$0.018 & 0.439$\pm$0.013 \\
& \checkmark & & &
& 0.674$\pm$0.009 & 0.503$\pm$0.009
& 0.642$\pm$0.015 & 0.446$\pm$0.012 \\
& \checkmark & \checkmark & &
& 0.693$\pm$0.011 & 0.524$\pm$0.020
& 0.647$\pm$0.009 & 0.451$\pm$0.013 \\
& \checkmark & \checkmark & \checkmark &
& 0.698$\pm$0.015 & 0.527$\pm$0.015
& 0.657$\pm$0.015 & 0.463$\pm$0.021 \\
& \checkmark & \checkmark & \checkmark & \checkmark
& 0.709$\pm$0.022 & 0.541$\pm$0.026
& 0.664$\pm$0.009 & 0.465$\pm$0.009 \\
\midrule
A$\beta$
& & & &
& 0.580$\pm$0.031 & 0.494$\pm$0.023
& 0.542$\pm$0.025 & 0.451$\pm$0.035 \\
& \checkmark & & &
& 0.606$\pm$0.037 & 0.553$\pm$0.055
& 0.550$\pm$0.045 & 0.461$\pm$0.046 \\
& \checkmark & \checkmark & &
& 0.626$\pm$0.035 & 0.572$\pm$0.044
& 0.588$\pm$0.028 & 0.508$\pm$0.008 \\
& \checkmark & \checkmark & \checkmark &
& 0.636$\pm$0.022 & 0.574$\pm$0.050
& 0.601$\pm$0.035 & 0.539$\pm$0.036 \\
& \checkmark & \checkmark & \checkmark & \checkmark
& 0.648$\pm$0.018 & 0.594$\pm$0.015
& 0.619$\pm$0.032 & 0.552$\pm$0.049 \\
\midrule
Tau
& & & &
& 0.640$\pm$0.032 & 0.395$\pm$0.055
& 0.559$\pm$0.040 & 0.340$\pm$0.031 \\
& \checkmark & & &
& 0.668$\pm$0.050 & 0.416$\pm$0.086
& 0.561$\pm$0.016 & 0.329$\pm$0.017 \\
& \checkmark & \checkmark & &
& 0.688$\pm$0.034 & 0.453$\pm$0.056
& 0.566$\pm$0.014 & 0.340$\pm$0.021 \\
& \checkmark & \checkmark & \checkmark &
& 0.704$\pm$0.029 & 0.470$\pm$0.058
& 0.569$\pm$0.014 & 0.342$\pm$0.023 \\
& \checkmark & \checkmark & \checkmark & \checkmark
& 0.725$\pm$0.018 & 0.493$\pm$0.047
& 0.570$\pm$0.018 & 0.343$\pm$0.020 \\
\midrule
Sex
& & & &
& 0.703\textpm0.040 & 0.656\textpm0.043
& 0.817\textpm0.013 & 0.790\textpm0.021 \\
& \checkmark & & &
& 0.792$\pm$0.014 & 0.755$\pm$0.017
& 0.837$\pm$0.013 & 0.808$\pm$0.028 \\
& \checkmark & \checkmark & &
& 0.806$\pm$0.013 & 0.780$\pm$0.017
& 0.842$\pm$0.017 & 0.813$\pm$0.028 \\
& \checkmark & \checkmark & \checkmark &
& 0.824$\pm$0.014 & 0.793$\pm$0.019
& 0.847$\pm$0.018 & 0.822$\pm$0.026 \\
& \checkmark & \checkmark & \checkmark & \checkmark
& 0.827\textpm0.011 & 0.798\textpm0.025
& 0.853\textpm0.015 & 0.830\textpm0.016 \\
\bottomrule
\end{tabularx}
\end{threeparttable}
\end{table}

\subsection{External validation}
To evaluate generalization across cohorts, we trained vSV-ViT using OASIS and evaluated the resulting models exclusively on subjects from the independent ADNI cohort. As shown in \cref{tab:external_oasis}, AUROC with CT for Dx, A$\beta$, tau, and sex was comparable to the pooled analysis (\cref{tab:main_results}). For Curv, AUROC was lower for Dx, A$\beta$, and sex, while remaining similar for tau. Overall, these results support the generalizability of vSV-ViT across cohorts.

\begin{table}[tb]
    \centering
    \setlength{\tabcolsep}{1.8pt}
    \renewcommand\bfdefault{b}
    \begin{threeparttable}
        \centering
        \small
        \caption{External validation performance of vSV-ViT trained on OASIS data and evaluated on the independent ADNI data. Diagnosis (Dx) is evaluated as multiclass classification of CN, MCI, and AD; A$\beta$ and tau are evaluated as binary classifications of PET-based positivity status; and sex is evaluated as binary classification. Cortical thickness and cortical curvature are evaluated separately as inputs. Values are reported as the mean $\pm$ standard deviation of the ADNI performance across the four OASIS cross-validation models; the ADNI cohort itself was not partitioned into folds.}
        \label{tab:external_oasis}
        \begin{tabularx}{\textwidth}{
        @{\hspace{3pt}}l
        >{\centering\arraybackslash}X
        >{\centering\arraybackslash}X
        >{\centering\arraybackslash}X
        >{\centering\arraybackslash}X
        }
            \toprule
            Target
            & \multicolumn{2}{c}{Cortical thickness}
            & \multicolumn{2}{c}{Cortical curvature} \\
            \cmidrule(lr){2-3}
            \cmidrule(lr){4-5}
            & AUROC & AUPRC & AUROC & AUPRC \\
            \midrule
            Dx & 0.715$\pm$0.008 & 0.536$\pm$0.018 & 0.613$\pm$0.018 & 0.412$\pm$0.011 \\
            A$\beta$ & 0.647$\pm$0.026 & 0.641$\pm$0.027 & 0.596$\pm$0.033 & 0.577$\pm$0.024 \\
            Tau & 0.727$\pm$0.021 & 0.377$\pm$0.032 & 0.572$\pm$0.014 & 0.258$\pm$0.022 \\
            Sex & 0.824$\pm$0.008 & 0.802$\pm$0.010
            & 0.825$\pm$0.006 & 0.794$\pm$0.004 \\
            \bottomrule
        \end{tabularx}
    \end{threeparttable}
\end{table}

\subsection{Limitations and Future Directions}
Although ADNI and OASIS are combined to improve cohort diversity, the analysis remains retrospective and depends on cohort-specific acquisition and biomarker-processing pipelines. PET-derived A$\beta$ and tau labels are not available for all MRI scans, producing task-specific sample sizes and class imbalance, especially for tau classification. As this study is a proof-of-concept for variable-size supervertices, vSV-ViT was validated only on Siemens 3T MRI data. Future work should evaluate the framework on multi-vendor datasets and extend it to additional surface-derived features or multimodal inputs. The current evaluation uses fsaverage6-resolution cortical surfaces; extension to higher-resolution meshes remains future work. Finally, future work will investigate the applicability of vSV-ViT to other 3D mesh learning tasks.

\section{Conclusion} 
In this study, we proposed variable-size supervertex (vSV) partitioning, which groups cortical surface vertices into vSV patches. Building on these vSVs, we designed vSV-ViT, which processes these patches through padding, mask-aware patch embedding, and vertex position embedding. Analysis of surface partitioning showed that vSV partitioning improved vertex assignment. Downstream validation demonstrated that vSV-ViT achieved higher predictive accuracy than recent models applicable to the cortical surface. These results support vSV-ViT as an effective framework for cortical surface learning.


\section*{Acknowledgments}
This work was supported by the National Research Foundation of Korea (NRF) grant funded by the Ministry of Science and ICT (MSIT) (RS-2024-00455720); the Advanced GPU Utilization Support Program and the High-Performance Computing Support project (RQT-25-070083), both funded by MSIT, Korea; the ANCHOR program funded by the Ministry of Education and Gyeonggi Province, Korea (2026-RISE-09-A37); and Hankuk University of Foreign Studies Research Fund of 2026. 

Data collection and sharing for this project was funded by the Alzheimer's Disease Neuroimaging Initiative (ADNI) (National Institutes of Health Grant U01 AG024904) and DOD ADNI (Department of Defense award number W81XWH-12-2-0012). ADNI is funded by the National Institute on Aging, the National Institute of Biomedical Imaging and Bioengineering, and through generous contributions from numerous private and corporate partners. A complete listing of ADNI investigators can be found at: \url{https://adni.loni.usc.edu/wp-content/uploads/how_to_apply/ADNI_Acknowledgement_List.pdf}.

Data were also provided in part by OASIS-3~\cite{lamontagne2019oasis3}, supported by NIH grants P30 AG066444, P01 AG03991, P01 AG026276, R01 AG043434, UL1 TR000448, and R24 HD008388.

\bibliographystyle{splncs04}
\bibliography{main}
\end{document}